\documentclass[letterpaper]{article}

\usepackage{ml2k}
\usepackage{tipa}
\usepackage{psfig}
\usepackage{epsf}
\usepackage[english]{babel}

\begin{document} 

\twocolumn[
\icmltitle{Meta-Learning for Phonemic Annotation of Corpora}    
 
\icmlauthor{V\'{e}ronique Hoste}{hoste@uia.ua.ac.be}
\icmlauthor{Walter Daelemans}{daelem@uia.ua.ac.be}
\icmlauthor{Erik Tjong Kim Sang}{erikt@uia.ua.ac.be}
\icmlauthor{Steven Gillis}{gillis@uia.ua.ac.be}
\icmladdress{CNTS Language Technology Group, University of Antwerp, Universiteitsplein 1, 2610 Wilrijk, Belgium}
\vskip 0.3in
]

{
\begin{picture}(0,0)
\put(20,185){
\makebox(440,0){
In:
{\em Proceedings of ICML-2000},
pages 375--382,
Stanford University, CA, USA, 2000.
}}
\end{picture}
}

\vspace*{-0.5cm}
\begin{abstract}
We apply rule induction, classifier combination and meta-learning
(stacked classifiers) to the problem of bootstrapping high accuracy
automatic annotation of corpora with pronunciation information.  The
task we address in this paper consists of generating phonemic
representations reflecting the Flemish and Dutch pronunciations of a
word on the basis of its orthographic representation (which in turn is
based on the actual speech recordings).  We compare several possible
approaches to achieve the text-to-pronunciation mapping task:
memory-based learning, transformation-based learning, rule induction,
maximum entropy modeling, combination of classifiers in stacked learning, and 
stacking of meta-learners. We are interested both in optimal accuracy and in
obtaining insight into the linguistic regularities involved.  
As far as accuracy is concerned, an already high accuracy level (93\% for
Celex and 86\% for Fonilex at word level) for single classifiers is 
boosted significantly with additional error reductions of 31\% and
38\% respectively using combination of classifiers, and a further 5\% 
using combination of meta-learners, bringing overall word level
accuracy to 96\% for the Dutch variant and 92\% for the
Flemish variant.  
We also show that
the application of machine learning methods indeed leads to increased
insight into the linguistic regularities determining the variation
between the two pronunciation variants studied.
\end{abstract}

\section{Introduction}
The context of this research is a large-scale Dutch-Flemish project
``Corpus Gesproken Nederlands" (Corpus Spoken Dutch, CGN) in which a 10
million word spoken corpus is collected and linguistically annotated.
One of the annotation layers is a representation of the pronunciation
of the recorded speech (the phonemic representation). As available
speech recognition technology is not yet up to generating this
annotation automatically, it has to be produced from the (manually
transcribed) orthographic transcription.  The task is complicated by
the fact that these pronunciation representations should reflect
either Flemish (the variant of Dutch spoken in the North of Belgium)
or Dutch pronunciation, depending on the origin of different parts of
the corpus.

To generate the phonemic representations, we need accurate
grapheme-to-phoneme conversion (the part of speech synthesis which
converts spelling into phonemic representations). It can be described
as a function mapping the spelling of words to their phonetic
symbols. Since the spelling of a word is ambiguous regarding its
pronunciation, what is a correct phonemic transcription is
contextually determined. One of the possibilities to tackle the
problem is to develop a system that captures the linguistic knowledge
of a given language in a set of rules with the disadvantage that hand
crafting linguistic rules is a rather difficult and time consuming
task. Moreover, this task has to be restarted every time a
grapheme-to-phoneme convertor is developed for a new
language. Examples of knowledge-rich expert systems are those of Allen
(1987) and of Divay and Vitale (1997).  Manual encoding of linguistic
information, however, is being challenged by data-driven methods since
the extraction of linguistic knowledge from a sample text corpus can
be a powerful method for overcoming the linguistic knowledge
acquisition bottleneck. Different approaches have already been used,
such as the use of learning algorithms to pronounce unknown words by
analogy to familiar lexical items (Dedina \& Nusbaum, 1991),
decision-tree learning (Dietterich, 1997), a neural network or
connectionist approach (Sejnowski \& Rosenberg, 1987) or memory-based
learning (Daelemans \& van den Bosch, 1996). Data-driven approaches
can yield comparable (and sometimes even more accurate) results than
the rule-based approach. 

In this paper, we are concerned with two questions. First, we
investigate the level of accuracy that can be obtained using various
machine learning techniques trained on two available lexical databases
containing examples of the pronunciation of Flemish and Dutch.  We
examine whether one variant of Dutch can add valuable information to
the prediction of the other in cascaded or stacked classifiers.
Different individual classifiers were combined in order to obtain an
improved estimator. In the machine learning literature, this approach
is called ensemble, stacked or combined classifiers (Dietterich,
1997). The underlying idea is that, when the errors
committed by the individual classifiers are uncorrelated to a
sufficient degree and their error rate is low enough, the resulting
combined classifier will perform better. This approach, while common
in the Machine Learning literature, has only recently been introduced
in natural language processing research (e.g., van Halteren,
Zavrel, and Daelemans (1998) for word class disambiguation). 
In the research presented here, we discuss the combination of different
classifiers trained on the same or slightly different tasks. This
contrasts with other ensemble methods which combine subsets of the
training data (as in bagging) or which combine multiple versions of the
training data in which previously misclassified examples get more weight
(as in boosting).
In this paper, we empirically examine whether these combined
classifiers result in substantial accuracy improvements in learning 
{\em grapheme-to-phoneme conversion}.
Our second research question concerns the use of rule induction as a
method to model the systematicity implicit in the differences between
Flemish and Dutch pronunciation.

The training data for our experiments consists of two lexical
databases representing Dutch and Flemish. For Dutch, Celex (release 2)
was used and for Flemish Fonilex (version 1.0b). The Celex database
contains frequency information (based on the INL corpus), and
phonological, morphological, and syntactic lexical information for
more than 384,000 word forms, and uses the DISC representation as
encoding scheme for word pronunciation. Fonilex is a list of
more than 200,000 word forms together with their Flemish
pronunciation. For each word form, an abstract lexical representation
is given, together with the concrete pronunciation of that word form
in three speech styles: highly formal speech, sloppy speech and
``normal'' speech (which is an intermediate level). A set of
phonological rewrite rules was used to deduce these concrete speech
styles from the abstract phonological form. The initial phonological
transcription was obtained by a grapheme-to-phoneme converter and was
afterwards corrected by hand. Fonilex uses YAPA (comparable to DISC)
as encoding scheme. The Fonilex entries also contain a reference to
the Celex entries, since Celex served as basis for the list of word
forms in Fonilex. The word forms in Celex with a frequency of 1 and
higher are included in Fonilex and from the list with frequency 0
(words not attested in a reference corpus), only the monomorphematic
words were selected.

In the following section, we first explain our experimental setup,
describing the data sets being used in the different experiments. An
overview of the experiments is also provided. In section 3, we
introduce the experimental methods and we go on to report
the overall results of the experiments. Section 4 shows that the use
of machine learning techniques, and especially the use of rule
induction techniques, leads to an increased insight into the
linguistic regularities determining the variation between the two
pronunciation variants. In a final section we conclude with a summary
of the most important observations.

\section{Experimental Setup}
The training data for the text-to-pronunciation experiments are two
corpora, representing the Northern Dutch and Flemish variants. 
The data set we used consists of all Fonilex entries with
omission of the double entries. In case of double word forms with
different possible transcriptions, all different transcriptions were
taken, as in the word ``caravan'', which can be phonemically
represented as
/\textipa{k}\textipa{A}\textipa{r}\textipa{A}\textipa{v}\textipa{A}\textipa{n}/
or as
/\textipa{k}\textipa{E}\textipa{r}\textipa{E}\textipa{v}\textipa{E}\textipa{n}/.
These double transcriptions only appear in Fonilex, which explains
why the text-to-pronunciation task for Flemish is more difficult. Also
words the phonemic transcription of which is longer than the
orthography and for which no compound phonemes are provided, are
omitted, e.g. "b'tje" (Eng.: ``little b'')(phonemically:
/\textipa{b}\textipa{e}\textlengthmark\textipa{t}\textipa{j}\textschwa/).
DISC is used as phonemic encoding scheme. All DISC phonemes are
included and new phonemes are created for the phonemic symbols which
only occur in the Fonilex data base.

Before passing the data through the machine learning program,
alignment (Daelemans \& van den Bosch, 1996) is performed for the
graphemic and phonemic representations of Celex and for those of
Fonilex, since the phonemic representation and the spelling of a word
often differ in length. Therefore, the phonemic symbols are aligned
with the graphemes of the written word form. In case the phonemic
transcription is shorter than the spelling, null phonemes (/-/) are
used to fill the gaps. In the example ``aalmoezenier'' (Eng.:
``chaplain'') this results in the following alignment:
\begin{table}[h]
\caption{The use of phonetic null insertion in the word ``aalmoezenier'' }
\vskip 0.15in
\begin{center}
\begin{small}
\begin{tabular}{|l|l|l|l|l|l|l|l|l|l|l|l|}
\hline
a & a & l & m & o & e & z & e & n & i & e & r \\
\hline
\textipa{a}\textlengthmark & - & \textipa{l} & \textipa{m} & \textipa{u}\textlengthmark & - & \textipa{z} & \textschwa & \textipa{n} & \textipa{i}\textlengthmark & - & \textipa{r} \\
\hline
\end{tabular}
\end{small}
\end{center}
\vskip -0.1in
\end{table}

A further step in the preparation of the data, consists of the use of
an extensive set of so-called ``compound phonemes''. Compound phonemes
are used whenever graphemes map with more than one phoneme, e.g. the
word 'jubileum' aligns to /j\}bIl]\}m/ in which the compound phoneme
/]/ stands for /ej/. Both alignment and the use of compound phonemes
leads to a corpus consisting of 173,874 word forms or 1,769,891
phonemes for each of the variants.

In order to achieve the grapheme-to-phoneme mapping task, we used
different approaches:
\begin{enumerate}
\vspace*{-0.30cm}
\item Training two single classifiers on lexical databases containing
 examples of the pronunciation of Dutch and Flemish, respectively,
 using memory-based learning.
\vspace*{-0.25cm}
\item Training classifiers for each pronunciation variant using the 
 predicted output for the other as an additional information source in
 \vspace*{-0.15cm}
 \begin{itemize}
 \item a cascaded approach and 
 \vspace*{-0.10cm}
 \item classifier combination.
 \end{itemize}
\vspace*{-0.25cm}
\item Trying to improve the results of classifier combination by
combining the combination classifiers.
\end{enumerate}
\vspace*{-0.30cm}

In these experiments, the text-to-pronunciation task is defined as the
conversion of fixed-size instances representing the grapheme with a
certain context to a class representing the target phoneme, as shown
in Table 2, using a technique proposed by Sejnowski and Rosenberg (1987).

\begin{table}[h]
\caption{Example of instances generated from the word ``eet''
(Eng. ``eat'') for the word-pronunciation task in the single classifier training experiment.}
\vskip 0.15in
\begin{center}
\begin{small}
\begin{tabular}{|l|l|l||l|}
\hline
left context 	& focus 	& right context 	& classification \\
= = = 		& e			& e t =			& e\\
= = e		& e			& t = =			& -\\
= e e		& t 			& = = = 		& t\\
\hline
\end{tabular}
\end{small}
\end{center}
\vskip -0.1in
\end{table}

In the
cascade and classifier combination experiments, the instances
contain both graphemic and phonemic information. In this study, we
choose a fixed window width of seven, which offers sufficient context
information for adequate performance. Extending the window would make
the meta-meta-classifier experiment computationally very costly. 

In all experiments, except when explicitly mentioned otherwise,
ten-fold cross-validation (Weiss \& Kulikowski, 1991) is used as
experimental method for error estimation. All experiments, both the
component and combination experiments were performed on the same data
set partitions for both variants of Dutch. E.g., in the classifier
combination experiment, where output of both a classifier trained on
Celex and a classifier trained on Fonilex is used as input, this
parallel way of working is necessary, since it has to be avoided that
one component classifier is trained on data held out in the training
of the other component classifier. A non-parallel way of working could
lead to over-optimistic accuracies for the classifier combination
experiments.

\section{Learning Dutch Word Pronunciation}
In the following subsections a brief introduction is given to each
approach, followed by a description of the experiments and a brief
discussion of the results.

\subsection{Single Classifiers}
In order to obtain a high accuracy grapheme-to-phoneme convertor,
different approaches were studied. In a first approach, one single
classifier is trained on Fonilex and another classifier on Celex.

For this experiment we have made use of Timbl (Daelemans et~al.,
1999), a software package implementing several memory-based learning
(lazy learning) techniques. {\em Memory-based learning} is a learning
method which is based on storing all examples of a task in memory and
then classifying new examples by similarity-based reasoning from this
memory of examples. The approach is argued to be especially suited for
natural language processing (NLP) because of the abundance of
sub-regularities and exceptions in most NLP problems (Daelemans, van
den Bosch, \& Zavrel, 1999), and has been successfully applied to the
grapheme-to-phoneme conversion problem before (Daelemans \& van den
Bosch, 1996).  The algorithm used for this experiment is called
IB1-IG. IB1-IG (Daelemans et~al., 1997) extends the basic k-nn
algorithm with information gain ratio (Quinlan, 1991) feature
weighting. IB1-IG builds a database of instances during
learning. During testing, the distance between a test item and each
memory item is defined as the number of features for which they have a
different value. IG (information gain ratio) weighting looks at each
feature in isolation and measures how much information it contributes
to the reduction of uncertainty about the correct class label. These
measures are used as feature weights in computing the distance between
items.

In Table 3, an overview is given of the generalisation accuracy using
the IB1-IG algorithm on Celex and Fonilex. For Celex, a
generalisation accuracy of 99.16\% is reached at the phoneme level,
and of 93.00\% on the word level. For Fonilex, which has a more complex
phonemic representation and in which word forms can have more than one
phonemic transcription, percentages are lower:
IB1-IG correctly classifies 98.18\% of the phonemes and 86.37\% of the
words.
\begin{table}[h]
\caption{Generalisation accuracy on the word and phoneme level of two single classifiers, trained on Celex and Fonilex, respectively. The last column provides the standard deviation on the phoneme level.}
\vskip 0.15in
\begin{center}
\begin{small}
\begin{tabular}{|p{1.5cm}|p{1.5cm}|p{1.5cm} p{1.5cm}|}\hline
               & Words           & Phonemes  &$\pm$ \\
\hline
Celex          & 93.00           & 99.16     & 0.03 \\
\hline
Fonilex        & 86.37           & 98.18     & 0.04 \\
\hline
\end{tabular}
\end{small}
\end{center}
\vskip -0.1in
\end{table}

Apart from the spelling, we did not have additional information to
further improve the generalization accuracy. Given that the
classifiers for Flemish and Dutch are trying to learn very similar but
nevertheless slightly different mappings, we investigate in the next
subsection whether the predicted output of the one could help in
making more accurate the predicted output of the other to further
improve the accuracy of the grapheme-to-phoneme convertors.

\subsection{Cascade and Classifier Combination}
In this section, the experiments in which single classifiers are
trained on Celex and Fonilex, respectively, are taken as the basis for
various experiments, as displayed in Figure 1. Four different
experiments are performed using this point of view. In all
experiments, the IB1-IG algorithm, as described in 3.1, is used to
perform the text-to-pronunciation mapping task.

\vspace*{-0.20cm}
\begin{itemize}
\vspace*{-0.20cm}
\item In (i) a single classifier is trained on one of both
pronunciation variants; in a second step,
the output of this process is used as input for training another
classifier for the other variant.
\vspace*{-0.20cm}
\item In (ii), the same information is used, but the spelling
information, together with the predicted output of the classifier
trained on the {\em other variant} in the experiment described in 3.1 is
used as an input pattern for a second classifier.
\vspace*{-0.20cm}
\item In (iii), spelling information together with the output of {\em both}
classifiers described in Section 3.1., is given to train a classifier.
\end{itemize}
\vspace*{-0.5cm}
\begin{figure}[h]
\vskip 0.2in
\begin{center}
\setlength{\epsfxsize}{2.75in}
\centerline{\epsfbox{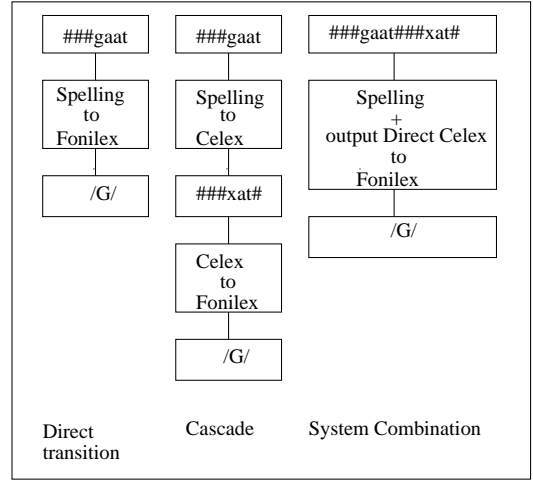}}
\caption{Architecture of three different approaches applied to the
Fonilex phonemic transcription: training a single classifier on
Fonilex, training a classifier for Fonilex using the predicted output
for Celex as an additional information source in a cascade and in classifier combination.} 
\end{center}
\vskip -0.2in
\end{figure} 
In Table 4, an overview is given of the generalisation accuracy of the
different classifiers. The combination classifier which generates the
highest percentage of generalisation errors is indicated in bold.
\begin{table}[h]
\caption{Generalisation accuracy of the cascaded approach and classifier combination.}
\vskip 0.15in
\begin{center}
\begin{small}
\begin{tabular}{|p{1.5cm}|p{1.5cm}|p{1.5cm} p{1.5cm}|}\hline
{\sc Celex}    & Words           & Phonemes    &$\pm$ \\
\hline
(i)            & 92.90           & 99.10       & 0.04\\
\hline
(ii)           & 94.18           & 99.28       & 0.03\\
\hline
(iii)          & {\bf 95.16}     & {\bf 99.40} & 0.02\\
\hline\hline
{\sc Fonilex}  &                 &             & \\
\hline
(i)            & 87.58           & 98.29       & 0.03 \\
\hline
(ii)           & 88.03           & 98.36       & 0.03\\
\hline
(iii)          & {\bf 91.55}     & {\bf 98.89} & 0.04  \\
\hline
\end{tabular}
\end{small}
\end{center}
\vskip -0.1in
\end{table}

For both Celex and Fonilex, the experiment in which spelling and
predicted output for both problems are combined in a meta-classifier
yields the highest accuracy: 99.40\% on the phoneme level for Celex
and 98.89\% for Fonilex, corresponding with 95.16\% and 91.55\%
respectively at the word level. Interestingly, adding a classifier in
the combination having learned a particular task (e.g. Flemish) can
help boost performance on a different but similar task (Dutch).

\subsection{Combining the Combination Classifiers}
In this section we further explore the use of system combination in
the grapheme-to-phoneme conversion task by combining combined
classifiers, as displayed in Figure 2. 

Four different meta-classifiers 
are used, viz. C5.O (described in
Section 4), IB1-IG (described in 3.1), IGTREE (Daelemans, van den
Bosch, \& Weijters, 1997) and MACCENT.\footnote{Details
on how to obtain Maccent can be found on: 
http://www.cs.kuleuven.ac.be/\~{ }ldh/} IGTREE is an optimised
approximation of the instance-based learning algorithm IB1-IG. In
IGTREE, the database of instances is compressed into a decision tree,
consisting of paths of connected nodes ending in leaves which contain
classification information. Information gain is used to
determine the order in which the feature values are added as arcs to
the tree. The last meta-classifier, MACCENT, is an implementation of 
maximum entropy modeling allowing symbolic features as input. The
package takes care of the
translation of symbolic values to binary feature vectors, and
implements the iterative scaling approach to finding the probabilistic
model. 
\begin{figure}[h]
\vskip 0.2in
\begin{center}
\setlength{\epsfxsize}{2.25in}
\centerline{\epsfbox{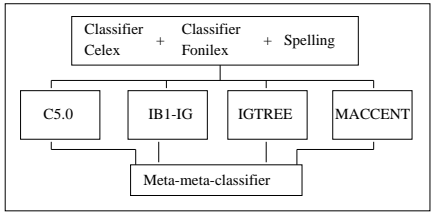}}
\caption{Architecture of the meta-meta-learning process. Component meta-classifiers are C5.O, IB1-IG, IGTREE, MACCENT}
\end{center}
\vskip -0.2in
\end{figure} 

The results of these four combination classifiers are used to train a
so-called ``meta-meta-classifier'', for the training of which IB1-IG
is used. The reasoning behind this experiment is that the same way a
meta-classifier can overcome some of the errors of different
``object-classifiers'' learning a similar task, a
``meta-meta-classifier'' should be able to do the same for
meta-classifiers. In these experiments, the combination classifiers
are trained on spelling information together with the output of both
object classifiers described in Section 3.1. The predictions of the
four stacked classifiers are then fed to a new combination classifier.

\begin{table}[h]
\caption{Generalisation accuracy of each meta-classifier and the
meta-meta-classifier (indicated by ``IB1-IG-meta'').} 
\vskip 0.15in
\begin{center}
\begin{small}
\begin{tabular}{|p{2.25cm}|p{1.25cm}|p{1.25cm} p{1.25cm}|}\hline
{\sc Celex} 	       & Words           & Phonemes    &$\pm$ \\
\hline	
(C5.0)  	       &   93.03         & 99.16       & 0.03 \\
\hline	
(IB1-IG)	       &   95.16         & 99.40       & 0.02\\
\hline	
(IGTREE)	       &   94.94         & 99.37       & 0.03\\
\hline
(MACCENT)	       &   92.07         & 99.03       & 0.05\\
\hline
(IB1-IG-meta)          &  {\bf 95.53}    & {\bf 99.45} & 0.02\\
\hline\hline
{\sc Fonilex}  	       &                 &             & \\
\hline
(C5.0)	               &   88.41         & 98.48       & 0.05 \\
\hline
(IB1-IG)      	       &   91.55         & 98.89       & 0.04\\
\hline
(IGTREE)       	       &   91.33         & 98.85       & 0.04\\
\hline
(MACCENT)	       &   87.27         & 98.28       & 0.04\\
\hline
(IB1-IG-meta)          &   {\bf 92.25}   & {\bf 98.99} & 0.03\\
\hline
\end{tabular}
\end{small}
\end{center}
\vskip -0.1in
\end{table}

In this section we reported research on the generation of a maximally
accurate phonemic representation for both Dutch and Flemish reflecting
the pronunciation of a given word on the basis of its orthographic
representation. In order to obtain high accuracy automatic annotation, 
different approaches were used. These
experiments showed that the memory-based learning algorithms
performed well on the text-to-pronunciation mapping task. 
Training single classifiers on both variants of Dutch already resulted in
generalisation accuracies of about 99\% at the phoneme level for Celex
and 98\% for Fonilex (93\% and 86\% at the word level respectively).  
Making use of classifier combination with
information predicted by a classifier for the other pronunciation
variant led to further reductions of the error at the word level of
about 31\% for Celex and 38\% for Fonilex to which meta-meta-learning 
added a limited but significant additional reduction (5\%).  
The already high accuracy level for single classifiers is boosted
significantly using combination of classifiers and combination of
meta-learners. With this high level of accuracy, automatic
phonemic conversion becomes an increasingly more useful annotation tool.

\section{Rule Induction}
Apart from being after high accuracy, we are also interested in
insight into the linguistic
regularities governing the differences between the two regional
variants of Dutch. Using rule induction techniques, we investigate
whether machine learning techniques reproduce the theoretical analysis
of linguists, and whether rules can be induced that accurately
translate one variant into the other.

\subsection{Experiments}
We first focus
on the question whether it is possible to predict one variant on the
basis of the phonemic representation of the other. Our starting point
is the assumption that the differences in the phonemic transcriptions
between Flemish and Dutch are highly systematic, and can be
represented in a set of rules, which provide linguistic
insight into the overlap and discrepancies between both
variants. Moreover, they can be used to adapt pronunciation databases
for Dutch automatically to Flemish and vice versa. 
In our experiment we used two rule induction techniques,
viz. Transformation-based error-driven learning (TBEDL) (Brill, 1995), a
learning method popular in NLP, and the well-known
C5.0 (Quinlan, 1993).

In {\em TBEDL}, transformation rules are learned by 
comparing a corpus that is
annotated by an initial-state annotator to a correctly annotated
corpus, which is called the ``truth''. In this study, the Fonilex
representation functions as ``truth'', and the Celex representation as
initial-state annotation. The task is to learn how to transform Celex
representations into Fonilex representations (i.e., translate Dutch
pronunciation to Flemish pronunciation). Rule induction is greedy, is
triggered by differences between the initial-state representations and
the truth, and constrained by a number of user-defined patterns
restricting the context. This learning
process results in an ordered list of transformation rules which
reflects the systematic differences between both representations. A
rule is read as: ``change x (Celex) into y 
(Fonilex) in the following triggering environment''. E.g., 
\vspace{-0.08in}
\begin{center}
\begin{tabular}{l}
/\textipa{i}\textlengthmark/ /\textipa{I}/ NEXT 1 OR 2 OR 3
PHON /\textipa{e}\textlengthmark/\\ 
(change a tense /i/ to a lax /i/ when one of the \\
three following Celex phonemes is a tense /e/). \\
\end{tabular}
\end{center}
\vspace{-0.08in}
{\em C5.0}, on the other hand, which is a commercial
version of the C4.5 program, generates a classifier in the form of a
decision tree. 
Since
decision trees can be hard to read, the decision tree is converted to
a set of production rules, which are more intelligible to the
user. The rules have the form ``L -$>$ R'', in which the left-hand
side is a conjunction of attribute-based tests and the right-hand side
is a class. When classifying a case, the list of rules is examined to
find the first rule whose left-hand side satisfies the case. In this
experiment we have made use of a context of three phonemes preceding
(indicated by f-1, f-2, and f-3) and following (f+1, f+2, f+3) the
focus phoneme, which is indicated by an 'f'. The predicted class for
this case is then the right-hand side of the rule. At the top of the
rule the number of training cases covered by the rule is given
together with the number of cases that do not belong to the class
predicted by the rule. The ``lift'' is the estimated accuracy of the
rule divided by the prior probability of the predicted class. E.g.,
\vspace{-0.08in}
\begin{center}
\begin{tabular}{l}
(6422/229, lift 79.0)\\
f = \textipa{i}\textlengthmark\\
f+1 in \{\textipa{m}, \textipa{b}, \textipa{t}, \textipa{r},
      \textipa{k}, \textipa{N}, \textipa{G}, \textipa{f}, \textipa{n},
    \textipa{v}, \textipa{h}, \textipa{d}, \textipa{l}, \textipa{p},\\
      \textipa{s}, \textipa{z}, \textipa{S}, (...)\}\\
 -$>$  class \textipa{I}  [0.964]\\
\end{tabular}
\end{center}
\vspace{-0.08in}
In TBEDL, the complete training set of 90\% was used for learning the
transformation rules. A threshold of 15 errors was specified, which
means that learning stops if the error reduction lies under that
threshold. For the C5.0 experiment, 50\% (796,841 cases) of the
original training set served as training set (more training data was
computationally not feasible on our hardware). A decision tree model
and a production rule model were built from the training cases. The
tree gave rise to 671 rules, which were applied to
the original 10\% test set we used in the Brill experiment. In order
to make the type of task comparable for the transformation based
approach used by TBEDL, in the classification-based approach used in
C5.0, the output class to be predicted by C5.0 was either `0' when the
Celex and Fonilex phoneme are identical (i.e. no change), or the
Fonilex phoneme when Celex and Fonilex differ (mimicking a
transformation approach). 

Table 6 gives an overview of the overlap between Celex and Fonilex
after application of both rule induction techniques. A comparison of
these results shows that, when evaluating both TBEDL and C5.0 on the
test set, the transformation rules learned by the Brill-tagger have a
higher error rate, even when C5.0 is only trained on half the data
used by TBEDL. On the word level, the initial overlap of 55.25\% 
is raised to 83.01\% after application of the 430 transformation rules,
and to 85.93\% when using the C5.0 rules. On the phoneme
level, the 92.20\% of the initial overlap is increased to 97.74\%
(TBEDL) and 98.14\% (C5.0). A closer analysis of the rules produced 
during TBEDL reveals that the first 50 rules lead to a
considerable increase of performance from 55.25\% to 76.19\% on the
word level and from 92.20\% to 96.62\% on the phoneme level, which
indicates the high applicability of these rules. Afterwards, the
increase of accuracy is more gradual: from 76.19\% to 83.01\% (words)
and from 96.62\% to 97.74\% (phonemes).

\begin{table}[h]
\caption{Overlap between Celex and Fonilex after application of all
transformation rules and C5.0 production rules.}
\label{sample-table}
\vskip 0.15in
\begin{center}
\begin{small}
\begin{tabular}{|p{1cm}|p{1.5cm}|p{1.5cm}|}
\hline
      &Words       & Phonemes     \\
\hline
TBEDL & 83.01      & 97.74     \\
\hline
C5.0  & 85.93      & 98.14     \\
\hline
\end{tabular}
\end{small}
\end{center}
\vskip -0.1in
\end{table}

When looking only at those cases where Celex and Fonilex differ, we
see that it is possible to learn transformation rules which predict
62.0\% of the differences at the word level and 71.0\% of the
differences at the phoneme level. The C5.0 rules are more or less 5-7\%
more accurate: 68.6\% (words) and 76.2\% (phonemes). It is indeed
possible to reliably `translate' Dutch into Flemish. These results,
however, are below the results generated in the preceding experiment
where there is a direct transition from spelling to Fonilex and from
spelling to Celex. The rule-induction process described above requires
a first component which does the transition from spelling to the
phonemic Celex transcription. In order to obtain
a Fonilex transcription, the rules generated by TBEDL or C5.0 are
applied to the output of the first component. 

\subsection{Linguistic Regularities}
In this section we will discuss some example rules generated for
consonants and vowels. Starting point is the first ten rules that were
learned during TBEDL, which will be compared with the ten C5.0 rules,
which most reduce the error rate.

\subsubsection{Consonants}
Nearly half of the differences on the consonant level concerns the alternation
between voiced and unvoiced consonants. In this group, the alternation
between /\textipa{x}/ and /\textipa{G}/ is the most frequent one.  In
the word ``gelijkaardig'' (Eng.: ``equal''), for example, we find a
/\textipa{x}\textschwa\textipa{l}\textipa{E}\textipa{i}\textipa{k}\textipa{a}\textlengthmark\textipa{r}\textipa{d}\textschwa\textipa{x}/
with a voiceless velar fricative in Dutch and
/\textipa{G}\textschwa\textipa{l}\textipa{E}\textipa{i}\textipa{k}\textipa{a}\textlengthmark\textipa{r}\textipa{d}\textschwa\textipa{x}/
with a voiced velar fricative in Flemish. The word ``machiavellisme''
(Eng.: ``Machiavellism'') is pronounced as
/\textipa{m}\textipa{A}\textipa{G}\textipa{i}\textlengthmark\textipa{j}\textipa{a}\textlengthmark\textipa{v}\textipa{E}\textipa{l}\textipa{I}\textipa{s}\textipa{m}\textschwa/
in Dutch and as
/\textipa{m}\textipa{A}\textipa{k}\textipa{I}\textipa{j}\textipa{A}\textipa{v}\textipa{E}\textipa{l}\textipa{I}\textipa{z}\textipa{m}\textschwa/
in Flemish. 

This alternation also is the subject of the first transformation rule
that was learned, namely ``\textipa{x} \textipa{G} PREV 1 OR 2 PHON
STAART'' which can be read as ``/\textipa{x}/ changes into
/\textipa{G}/ in case of a word beginning one or two positions
before''. When looking at the ten most important C5.0 rules, 
this alternation is described in:
\vspace{-0.08in}
\begin{center}
\begin{tabular}{l}
(6814/27, lift 109.5)\\
f-1 in \{=, \textipa{E}\textlengthmark\}\\
f = \textipa{x}\\
-$>$  class \textipa{G}  [0.996]\\
\end{tabular} 
\end{center}
\vspace{-0.08in}
Another important phenomenon is the use of palatalisation in Flemish,
as in the word ``aaitje'' (Eng.: ``stroke''), where Fonilex uses the
palatalized form
/\textipa{a}\textlengthmark\textipa{j}\textipa{t}\textipa{S}\textschwa/
instead of
/\textipa{a}\textlengthmark\textipa{j}\textipa{t}\textipa{j}\textschwa/. 
This change is also described by both top ten Brill and C5.0 rules.\\

\subsubsection{Vowels}
The most frequent difference at the vowel level between 
Dutch and Flemish concerns the use of a lax vowel instead of a tense 
vowel for the
/\textipa{i}\textlengthmark/, /\textipa{e}\textlengthmark/,
/\textipa{a}\textlengthmark/, /\textipa{o}\textlengthmark/ and
/\textipa{u}\textlengthmark/. Tense Celex-vowels not only correspond
with tense, but also with lax vowels in Fonilex. Other less frequent
differences are glide insertion, e.g. in ``geshaket'' and the use of
schwa instead of another vowel, as in ``teleprocessing'' in Flemish.

Five out of the first ten transformation rules indicate a transition from
a tense vowel into a lax vowel in a certain triggering environment. A
closer look at the top ten C5.0 production rules shows that
seven rules describe this transition from a Celex tense
vowel to a Fonilex lax vowel. An example is the word ``multipliceer''
(Eng.: ``multiply'') which is transcribed as
/\textipa{m}\textbaru\textipa{l}\textipa{t}\textipa{i}\textlengthmark\textipa{p}\textipa{l}\textipa{i}\textlengthmark\textipa{s}\textipa{e}\textlengthmark\textipa{r}/
in Celex and as
/\textipa{m}\textbaru\textipa{l}\textipa{t}\textipa{I}\textipa{p}\textipa{l}\textipa{I}\textipa{s}\textipa{e}\textlengthmark\textipa{r}/
in Fonilex. The change of the second /\textipa{i}\textlengthmark/ into
a /\textipa{I}/ is described in the following transformation rule:
``/\textipa{i}\textlengthmark/ changes into /\textipa{I}/ if the NEXT
1 OR 2 OR 3 PHON is an /\textipa{e}\textlengthmark/.  The
corresponding C5.0 rule describing this phenomenon is the
following:
\vspace{-0.08in}
\begin{center}
\begin{tabular}{l}
         (7758/623, lift 75.4)\\
	 f = \textipa{i}\textlengthmark\\
	 f+1 in \{\textipa{m}, \textipa{b}, \textipa{t},
	 \textipa{k}, \textipa{G}, \textipa{f}, \textipa{n},
	 \textipa{v}, \textipa{d}, \textipa{p}, \textipa{s},
	 \textcommatailz, \textipa{g}\}\\
	 f+2 in \{\textipa{m}, \textipa{O}, \textipa{t},
	 \textipa{r}, \textipa{k}, \textipa{y}, \textipa{A},
	 \textipa{f}, \textipa{i}\textlengthmark,
	 \textipa{e}\textlengthmark, \textipa{n}, (...)\}\\
	 -$>$  class \textipa{I}  [0.920]\\
\end{tabular}
\end{center}
\vspace{-0.08in}
These rules, describing the differences between Dutch and
Flemish consonants and vowels also make linguistic sense. Linguistic
literature, such as (Booij, 1995) indicates tendencies such as voicing
and devoicing on the consonant level and the confusion of tense and
lax vowels as important differences between Dutch and Flemish. The 
same discrepancies are found in the transcriptions made
by Flemish subjects in the Dutch transcription experiments described 
in Gillis (1999).

\section{Conclusion and Future Work}
The development of accurate and understandable annotation tools is of
prime importance in current Natural Language Processing research,
which is based to a large extent on the development of reliable
corpora. We discussed the task of phonemic annotation in such a
large-scale corpus development project. We were able to show that for
this text-to-pronunciation task, machine learning techniques provide 
an excellent approach to bootstrapping the annotation and modeling 
the linguistic knowledge involved. We do not know of any approach 
based on hand-crafting with similar or better accuracy for 
grapheme-to-phoneme conversion for Dutch.

We were both interested in optimal accuracy and in obtaining increased
insight into the linguistic regularities involved. We have empirically
examined whether combination of different systems (in this case 
classifiers trained on different variants of Dutch) enables 
us to raise the performance
ceiling which can be observed when using data driven systems. A
comparison of the results of training single classifiers and the use
of a meta-classifier indeed indicates a significant decrease in error of
31\% Dutch and 38\% for Flemish. Going one step further, namely
combining the combination classifiers results in an additional error
decrease of 5\% for both Flemish and Dutch.

The use of rule induction techniques to predict one variant on the
basis of the phonemic transcription of the other variant, on the other
hand, generates more generalisation errors. However, This rule
induction process leads to an increased insight into the systematic
differences between both variants of Dutch.

In the text-to-pronunciation task, described in this study,
disambiguation in context is required, which is also the case for
other problems in language processing, such as tagging and chunking.
Therefore, we plan to explore whether combining classifiers and
combining combined classifiers can lead to accuracy boosts for these
other NLP problems as well. We will also investigate other methods
that have proved to be promising combination methods for our task (e.g. Naive
Bayes). A possible limitation of the current
approach may be that different tasks can only be combined when they
are very similar (in this case pronunciation prediction of two related
dialects), a situation which may be rare.

\section*{Acknowledgements} 

This research is partially supported by the ``Linguaduct'' project
(FWO Flanders, G.0157.97), and the project ``Spoken Dutch Corpus
(CGN)'' (Dutch Organization for Scientific Research NWO and the
Flemish Government).

\section*{References}
{\parindent -10pt\leftskip 10pt

Allen, J., Hunnicutt, S., \& Klatt, D. (1987). {\it From text to
speech: The MITalk sytem\/}. Cambridge: Cambridge University Press.

Booij, G. (1995). {\it The phonology of Dutch\/}. Oxford: Clarendon Press.

Brill, E. (1995). Transformation-based error-driven learning and
natural language processing: A case study in part of speech
tagging. {\it Computational Linguistics\/}, {\it 21\/}, 543-565.

Daelemans, W., van den Bosch, A., \& Weijters, T. (1997). IGTree: Using
trees for compression and classification in lazy learning
algorithms. {\it Artificial Intelligence Review\/}, {\it 11\/}, 407-423.

Daelemans, W., van den Bosch, A., \& Zavrel, J. (1999). Forgetting
exceptions is harmful in language learning. {\it Machine Learning}, 
{\it 34\/}, 11-43.

Daelemans, W., Zavrel, J., van der Sloot, K., \& van den Bosch,
A. (1999). {\it TiMBL: Tilburg memory based learner version 2.0
reference guide\/} (Technical Report-ILK 99-01). Induction of
Linguistic Knowledge Research Team, Tilburg. 

Daelemans, W., \& van den Bosch, A. (1996). Language-independent
data-oriented grapheme-to-phoneme conversion. In Van Santen, J.,
Sproat, R., Olive, J. \& Hirschberg, J. (Eds.), {\it Progress in speech
synthesis\/}. New York: Springer Verlag. 

Dedina, M.J., \& Nusbaum, H.C. (1991). PRONOUNCE: A program for
pronunciation by analogy. {\it Computer Speech and Language\/},
{\it 5\/}, 55-64.

Dietterich, T.G. (1997). Machine learning research: Four current
directions. {\it AI Magazine\/}, {\it 18\/}, 97-136.

Divay, M., \& Vitale, A.J. (1997). Algorithms for grapheme-phoneme
translation for English and French: Applications. {\it
Computational Linguistics\/}, {\it 23\/}, 495-523.

Gillis, S. (1999). {\it Phonemic transcriptions: Qualitative and
quantitative aspects\/} (Unpublished manuscript). CNTS Language
Technology Group, Antwerp. 

Quinlan, J.R. (1993). {\it C4.5: Programs for machine learning\/}. San
Mateo: Morgan Kaufmann Publishers.

Roche, E., \& Schabes, Y. (1995). Deterministic part-of-speech tagging
with finite-state transducers. {\it Computational Linguistics\/},
{\it 21\/}, 227-253.

Sejnowski, T.J., \& Rosenberg C.S. (1987). Parallel networks that
learn to pronounce English text. {\it Complex Systems\/},
{\it 1\/}, 145-168.

Van den Bosch, A., \& Daelemans, W. (1993). Data-Oriented methods for
grapheme-to-phoneme conversion. {\it Proceedings of the European
Chapter of the Association for Computational Linguistics\/} 
(pp. 77-90). Utrecht: Association for Computational Linguistics.

Van Halteren, H., Zavrel J., \& Daelemans, W. (1998). Improving data
driven wordclass tagging by system combination. {\it Proceedings of
the Joint Seventeenth International Conference on Computational Linguistics
and Thirty-sixth Annual Meeting of the Association for Computational
Linguistics\/} (pp. 491-497). Montreal: Association for Computational 
Linguistics.

Weiss, S., \& Kulikowski, C. (1991). {\it Computer systems that
learn\/}. San Mateo, CA: Springer Verlag.

}

\end{document}